\documentclass[sigconf]{acmart}

\usepackage{booktabs} 
\usepackage[english]{babel}
\usepackage[utf8x]{inputenc}
\usepackage[T1]{fontenc}
\usepackage{amsmath}
\usepackage{graphicx}
\graphicspath{ {files/} }
\usepackage{subfigure}

\usepackage[colorinlistoftodos]{todonotes}

\copyrightyear{2018}
\acmYear{2018}
\setcopyright{acmlicensed}
\acmConference[Multimedia for RETech'18]{Workshop on Multimedia
for Real Estate Tech}{June 11, 2018}{Yokohama, Japan}
\acmBooktitle{Multimedia for RETech'18: Workshop on Multimedia for Real Estate Tech, June 11, 2018, Yokohama, Japan}
\acmPrice{15.00}
\acmDOI{10.1145/3210499.3210526}
\acmISBN{978-1-4503-5797-5/18/06}

\begin{document}
\title{Visual Estimation of Building Condition with Patch-level ConvNets}

\author{David Koch}

\affiliation{%
  \institution{Kufstein University of Applied Sciences}
}
\email{david.koch@fh-kufstein.ac.at}

\author{Miroslav Despotovic}
\affiliation{%
  \institution{Kufstein University of Applied Sciences}
}
\email{miroslav.despotovic@fh-kufstein.ac.at}

\author{Muntaha Sakeena}
\affiliation{%
  \institution{St.P\"olten University of Applied Sciences}
}
\email{msakeena@fh-kufstein.ac.at}

\author{Mario Döller}
\affiliation{%
  \institution{Kufstein University of Applied Sciences}
}
\email{mario.doeller@fh-kufstein.ac.at}

\author{Matthias Zeppelzauer}
\affiliation{%
  \institution{St.P\"olten University of Applied Sciences}
}
\email{m.zeppelzauerh@fhstp.ac.at}

\begin{abstract}
The condition of a building is an important factor for real estate valuation. Currently, the estimation of condition is determined by real estate appraisers which makes it subjective to a certain degree. We propose a novel vision-based approach for the assessment of the building condition from exterior views of the building. To this end, we develop a multi-scale patch-based pattern extraction approach and combine it with convolutional neural networks to estimate building condition from visual clues. Our evaluation shows that visually estimated  building condition can serve as a proxy for condition estimates by appraisers. 

\end{abstract}



\maketitle

\section{Introduction}

A house is made up of many characteristics, all of which may affect its value. Hedonic regression analysis is typically used to estimate the marginal contribution of these characteristics and to predict real estate prices \cite{Sirmans2005}. There exist numerous studies on hedonic pricing in theoretical and empirical work, see for example \cite{Sirmans2005, Sirmans2010, Malpezzi2003}. In general the hedonic price function takes the form
\begin{equation}
\begin{split}
&  P_{i} = f(S_{i}, L_{i}, N_{i})\
\end{split}
\label{eq:quantil.description}
\end{equation}
where P$_{i}$ is the logarithm of the price or rent of house $i$, S$_{i}$ is a vector of structural housing characteristics, L$_{i}$ is a vector of location variables and N$_{i}$ is the neighborhood characteristics. In this paper, we focus on structural housing characteristics (e.g. square footage, number of bathrooms, age) and in particular the  \emph{condition} of a building with the aim to assess building condition automatically. The main research question of our work is: \textit{can the condition of the building be estimated reliably in an automated fashion from an unconstrained photograph of the building by computer vision algorithms}? This focus is motivated by the following two observations: Firstly, the condition of a property has a significant influence on its value, which is reflected also in hedonic price models, see for example \cite{Herath2015, Koch2015}. There are a number of studies, particularly in connection with age and depreciation, see \cite{Baum1993, Knight1996, Allen1997, Harrison2004, Iwata2008, Carter2011, manganelli2013, Zahirovich-Herbert2014, He2016}, 
which show that the condition of a building is an important factor for real estate valuation. Secondly, the condition is usually assessed by appraisers or brokers subjectively. This makes condition differ substantially from other variables like, e.g. year of construction, which can usually be estimated exactly and do not offer room for interpretation. The same applies to the variables like the presence of a balcony and the number of rooms, etc. The condition, however, is a variable that is not clearly and objectively defined and that often has a subjective bias. We hypothesize that real estate image analysis (REIA) is a promising means to capture the condition of a building in a more standardized and objective way.

Based on the work in \cite{ZeppelzauerICMR}, we present an approach for the estimation of building condition from image patches. Experiments with a large dataset show that useful visual clues for the estimation of condition can be extracted automatically and that the estimated condition has a positive impact on price estimation.

\section{Related Work}

Real estate prices can only be estimated based on a proper valuation model and highly depends on the selection of characteristics which are likely to reflect the real value. Price estimation usually takes building parameters in the form of readily available metadata into account, such as building age, number of stories and number of units \cite{BCI_2008}. Rich literature exists that focuses on price estimation based on metadata \cite{Michal_2015,Michal_2016,Michal_2017}.  

Artificial neural networks are capable of learning complex regression and classification models based on labeled training data and have shown to generalize well to unseen data. Thus, they have been increasingly used for price estimation of properties based on building metadata in the past and provide a satisfactory proxy for hedonic models \cite{ann_ts1, houseprice}. See \cite{comparison} for a comparison between hedonic models and artificial neural networks. Neural networks have further shown to be extremely powerful for the extraction of visual features and image classification \cite{krizhevsky2012imagenet}. Only little work has been done so far on leveraging visual information for real estate evaluations (real estate image analysis, REIA). First  REIA approaches include Eman et al. \cite{Ahmed2016HousePE} who combine visual features extracted from photographs with metadata provided by real estate companies. They extract SURF features \cite{bay2008speeded} from indoor and outdoor pictures of buildings, combine them with traditional building parameters and train a regression-based neural network from these inputs for price estimation. 
Omid et al. \cite{DBLP:journals/corr/PoursaeedMB17} analyze interior and exterior photos to assess the luxury level of a building and further use this estimate as additional parameter in traditional house price estimation. In comparison with Eman et al. they used a much larger dataset with multiple interior and exterior photos of specific houses.  Despotovic et al. \cite{Despotovic2018} leveraged visual building features derived from outdoor building pictures to predict the approximate heating energy demand of a building by using convolutional neural networks (CNNs). 
 
Our goal is to predict the condition of a house automatically from an unconstrained exterior view. This is based on the assumption that the actual condition of a building is reflected in its visual appearance (e.g. impurities of the facade, state of windows, etc.). Contrary to previous works, we employ only a single exterior image for our analysis and no interior images (often not available in practice). This challenges the analysis because weather and lighting conditions strongly influence the visual appearance. Furthermore, instead of predicting the price directly, we assess the relative \emph{discount} of price given by the building condition. Thus, the actual real estate prices (also difficult to get in practice) are not necessary for our method.

\section{Approach}
\label{sec:approach}

\begin{figure}[t]
\includegraphics[width=0.47\textwidth]{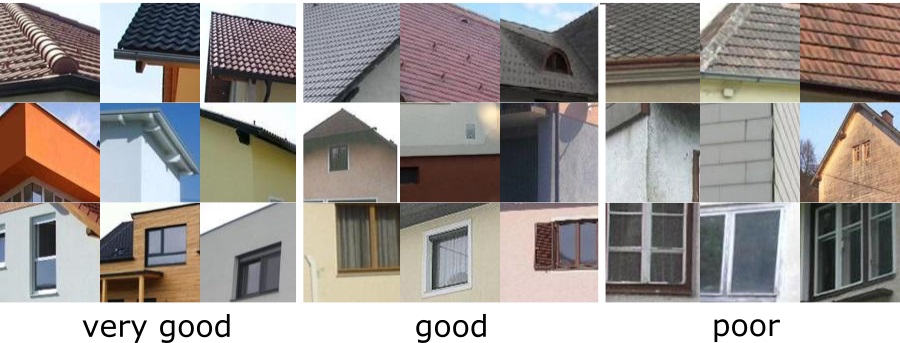}
\caption{Visual appearance of the building elements roof (top row), facade (middle row), and window (bottom row) for the conditions very good, good, and poor.}
\label{fig:facadecondition}
\end{figure}

We assume that the condition of a building can be detected by means of various visual clues, such as uneven structure and cracks in the facade, damaged painting of doors and windows, weathering of the roof, etc. We have a priori carried out a qualitative analysis of building images in order to examine which visual features potentially relate to the condition of a building. To this end, we have studied building images in detail to determine representative patterns. Figure \ref{fig:facadecondition} shows patches representing three different condition classes (excellent, good, poor) for different building elements (e.g. roofs and windows). We observe that aside from the facade of the building architectural elements, such as windows, doors, stairs, plinths, and roofs have potential to indicate its condition. 
The identified clues are rather at a local than a global level of a building. We thus design our approach to analyze the images locally in a patch-wise manner. 
We first extract patches from the images in a dense manner and then select those which are most likely to contain meaningful structure (Section \ref{subsec:patchExtraction}). Next, we train a CNN from the models that predicts different condition categories from the individual patches. Finally, to obtain an estimate for an entire building, we classify all its patches and perform majority voting on the patch-level predictions (Section \ref{subsec:conditionPrediction}).

\subsection{Patch Extraction and Selection}
\label{subsec:patchExtraction}

The input to our approach are unconstrained images of the exterior of a building. The buildings may appear in different scales, perspectives, lighting condition, and parts of the buildings may be cropped. In a first step, we extract patches from the images. We apply regular sampling (as in Dense Scale Invariant Feature Transform, DSIFT \cite{lowe1999object}). At each location we extract overlapping patches of different scales to account for scale variations, similarly to PHOW features \cite{bosch2007image}. This scheme results in a large set of partially redundant patches.

To reduce the number of patches and to increase the heterogeneity of captured visual patterns, we apply the following patch selection strategy. We extract the SIFT descriptor for each patch and apply k-means clustering for all patches of a given input image. From the resulting $k$ clusters we extract the patch which is closest to the respective cluster centroid as a representative of the cluster. Only these $k$ representative patches are considered further. To remove patches with low information content, i.e. low contrast (e.g. in areas of sky, roads) we take the norm of the SIFT descriptor as an indicator. Since the SIFT descriptor represents a gradient histogram, descriptors with small norm contain less edge-like structures than descriptors with larger norm. We select a fixed percentage $t$ of those representative patches with the largest norm for further processing.

The resulting set of patches capture heterogeneous patterns with high contrast. Using patches with multiple scales facilitates capturing building elements of different sizes. Overlap between the patches increases the spatial coverage of the facades. Patch extraction shows that most parts of the facades are captures well and densely. Homogeneous regions (e.g. sky but also homogeneous regions on the facade or on the roof) are widely omitted. Nevertheless, patterns not related to buildings may still be contained in this set, such as parts of cars, traffic signs, trees and bushes. To better filter out such irrelevant patches, we train a classifier that categorizes patches into 12 different classes of non-related objects and one class of building-related patterns. Non-related objects include e.g. cars, trees, people, asphalt, and poles. For classification we employ a Convolutional Neural Network (CNN), i.e., Alexnet \cite{krizhevsky2012imagenet} which was pre-trained on ImageNet. We modify the output layer to fit our target classes and fine tune the network with a training dataset containing 8000 labeled patches. The fine tuned network is applied to all candidate patches extracted by our approach.

\subsection{Building Condition Prediction}
\label{subsec:conditionPrediction}

The prediction of building condition is a two-stage approach. In the first stage we classify each input patch that passes the patch selection from Section \ref{subsec:patchExtraction} into a set of pre-defined condition classes. For this purpose, we train a network to classify individual patches. We employ residual networks (ResNet50 \cite{He2016CVPR}) due to their lower number of parameters (as e.g. compared to AlexNet) to speed up training. We adapt the output layer to the building condition classes and iteratively fine tune all layers of the network. To increase the heterogeneity of the training patches, we apply data augmentation (e.g. flipping, cropping, brightness variations) to the patches. 

In the second stage, we infer the condition of an entire building from the predictions at patch level. First, each patch extracted for a building is classified into a distinct condition category. Second, all patch-wise predictions are aggregated to obtain a final prediction for the entire building. Aggregation is either performed by majority voting (MV) on the patch-wise predictions or by averaging the class likelihoods (LH), i.e. the outputs of the softmax layer of the individual patches. To make aggregation more selective we filter out ambiguous patches, i.e. patches for which the class likelihoods do not show a clear winner. A patch is considered ambiguous if the difference between the highest and the second highest class likelihood is below a certain threshold (0.25 in our experiments).

\section{Experimental Setup}

\subsection{Dataset}
\label{subsec:dataset}
For our experiments, we use data provided by a renowned Austrian real estate software provider. The dataset consists of RGB images of single-family houses with different resolutions and qualities as well as JSON files containing house characteristics (metadata)\footnote{The images used in this work are not under creative commons license and thus cannot be shared, we will, however, publicly share extracted features and trained networks under \url{https://phaidra.fhstp.ac.at/detail_object/o:2960}.}. 
The images show the exterior of the houses and were taken from different angles, distances, and perspectives by different real estate experts. Weather and lighting conditions vary across the images. 
The available metadata include among others condition scores for all buildings ranging from c1 (best) to c9 (worst). 

The first two categories c1 and c2 refer to buildings that are as good as new and free of defects. Category c3 and c4 means normal conditions, i.e. only usual maintenance works are necessary. Categories from c5 onwards refer to buildings that need different amounts of repairs. 
Due to the semantic similarity of the above mentioned classes and the imbalanced class cardinalities (see also \ref{tab:initialClasses}) we group the nine categories into three classes: ``A'' (good condition), ``B'' (normal condition), and ``C'' (needs repairs) and use these three condition classes for our experiments. 
 
We partition the dataset randomly into a training, validation, and test set. Table \ref{tab:initialClasses} shows the distribution of the different partitions and categories as well as the aggregation of condition categories to the three target classes (``A'', ``B'', and ``C''). Table \ref{tab:uniqueHouses} shows the number of individual houses for each partition and target class of the dataset. Although for some houses more than one image exists, all images are independently processed and assessed. 

\begin{table}[t]
\begin{center}
\resizebox{0.9\linewidth}{!}{
    \begin{tabular}{c | r | r | r | c}
    Category & Training & Validation & Test & Target Class \\ \hline
    c1 & 2416 & 387 & 702 & A \\ 
    c2 & 83 & 24 & 35 & A \\ \hline 
    c3 & 3212 & 433 & 975 & B \\ 
    c4 & 121 & 23 & 33 & B \\ \hline	
    c5 & 1298 & 184 & 355 & C \\ 
    c6 & 102 & 15 & 34 &  C \\ 
    c7 & 120 & 15 & 44 & C \\ 
    c8 & 2 & 2 & 2 & C \\ 
    c9 & 13 & 3 & 3 & C \\ 
    \end{tabular}
    }
\end{center}
    \caption{Employed dataset: distribution of images across the three partitions of the dataset (train, validation, test) and the condition categories. The last column shows the aggregation of condition categories to target classes for classification.}
	\label{tab:initialClasses}
\end{table}

\begin{table}[t]
\begin{center}
\resizebox{0.9\linewidth}{!}{
    \begin{tabular}{l | c | r | r}
    Partition & Target Class & Unique Houses & Available Images \\ \hline
     & A & 1272 & 2499 \\ 
   Training & B & 1931 & 3333 \\ 
    & C & 1040 & 1535 \\ \hline
     & A & 206 & 411 \\ 
  Validation  & B & 276 & 456 \\ 
    & C & 205 & 219 \\ \hline
     & A & 376 & 737 \\ 
  Test  & B & 576 & 1008 \\ 
    & C & 391 & 438 \\ \hline	
    \textbf{Sum} &  & \textbf{6273} & \textbf{10636} \\

    \end{tabular}
    }
\end{center}
    \caption{Dataset characteristics: number of unique houses for each partition and target class of the dataset.}
	\label{tab:uniqueHouses}
\end{table}

\subsection{Setup \& Training}
For training, we extract multiple patches of different sizes from the images in the training set and select the $t=21\%$ highest-contrast patches from $k=50$ patch clusters, obtained by the patch selection strategy from Section \ref{subsec:patchExtraction}. These parameters were found to represent a good tradeoff between the amount of resulting data, redundancy in the patches and spatial coverage. Experiments with smaller datasets (lower percentage, less clusters) lead to weaker results. Larger datasets were not tested so far due to computational costs. For classification, we adapt the ResNet50 model (pre-trained on ImageNet) \cite{He2016CVPR} and shrink the output layer to three neurons, referring to classes ``A'', ``B'' and ``C'' (see Section \ref{subsec:dataset}). From the training patches we re-train all layers of the ResNet50 model. Training is performed for 30 epochs with a learning rate of 0.0001, a momentum of 0.9 and a decay of 0.0005. We apply a comprehensive data augmentation on all patches including cropping, flipping, scaling as well as brightness and color transformations. The total number of training patches is 153,356. The same procedure is applied to the validation images resulting in a set of 21,995 validation patches. We implemented the approach in Matlab and used the MatCovNet framework \cite{vedaldi2015matconvnet} for training the network. The experiments have been performed on a workstation with Ubuntu 14 OS, 64 GB RAM and an NVIDIA GTX 1080Ti.

\subsection{Evaluation \& Research Questions}

Our evaluation comprises three basic research questions:
\begin{enumerate}
  \item How accurately can the three conditions captured by the target classes (``A'': good condition, ``B'': normal condition, and ``C'': needs repairs) be differentiated automatically?
  \item Does the network learn meaningful visual patterns that correlate with the visually assessable condition?
  \item How does the automatically extracted building condition compare to the condition provided by experts in the prediction of cost-related parameters?
\end{enumerate}

To account for the first question, we compute the classification accuracy and analyze the classification confusions in Section \ref{subsec:prediction}. To investigate the second question, we investigate patches classified with different confidences by the network to analyze on which visual information its decisions are based on (see Section \ref{subsec:qualitative}). To investigate the third question, we design a regression model to predict the discount of the building (which is closely related to its cost) and compare regression performance between the model based on automatically extracted condition and the model based on expert assessments (see Section \ref{subsec:cost prediction}).

\section{Results}

\subsection{Classification Results}
\label{subsec:prediction}
 We compute the confusion matrix and classification accuracy to examine the performance of our approach in predicting building condition. Table \ref{tab:confusion matrix} shows the confusion matrices based on majority voting (MV) and average likelihood (LH), see Section \ref{subsec:conditionPrediction} for the three target classes. Confusion matrices contain the true labels along the vertical axis and the predicted labels along horizontal axis. Along the diagonal are the correct predictions (true positives) while off-diagonal elements show the incorrect predictions\footnote{Note that the total number of images per class may vary from the numbers in Table \ref{tab:initialClasses} and \ref{tab:uniqueHouses} because in some cases patch filtering (see Section \ref{subsec:patchExtraction}) removes all patches from a given input image. Such images are not included in the confusion matrices.}.
 
 \begin{table}[ t]

\setlength\unitlength{0.75cm}
\begin{picture}(4,4)
\multiput(0.1,0.1)(0,1){4}{\line(1,0){3}}
\multiput(0.1,0.1)(1,0){4}{\line(0,1){3}}
\put(0.4,0.5){67}
\put(1.3,0.5){163}
\put(2.3,0.5){206}

\put(0.3,1.5){227}
\put(1.3,1.5){713}
\put(2.5,1.5){67}

\put(0.3,2.5){505}
\put(1.3,2.5){205}
\put(2.5,2.5){25}

\put(-0.3,2.5){A}
\put(-0.3,1.5){B}
\put(-0.3,0.5){C}

\put(0.5,3.2){A}
\put(1.5,3.2){B}
\put(2.5,3.2){C}
\put (-0.8,1.3){\rotatebox{90}{True}}
\put(1.1,3.7){Predicted}
\end{picture}
\quad\begin{picture}(4,4)
\multiput(0.1,0.1)(0,1){4}{\line(1,0){3}}
\multiput(0.1,0.1)(1,0){4}{\line(0,1){3}}
\put(0.4,0.5){64}
\put(1.3,0.5){166}
\put(2.3,0.5){206}

\put(0.3,1.5){214}
\put(1.3,1.5){711}
\put(2.5,1.5){82}

\put(0.3,2.5){491}
\put(1.3,2.5){211}
\put(2.5,2.5){33}

\put(-0.3,2.5){A}
\put(-0.3,1.5){B}
\put(-0.3,0.5){C}

\put(0.5,3.2){A}
\put(1.5,3.2){B}
\put(2.5,3.2){C}

\put (-0.8,1.3){\rotatebox{90}{True}}
\put(1.1,3.7){Predicted}
 
\end{picture}

    \caption{Test results: confusion matrices showing true positive and false classifications on the test images with left: majority voting (MV), right: average likelihood (LH).}
\label{tab:confusion matrix}
\end{table}
Results from Table \ref{tab:confusion matrix} show that MV is the slightly better aggregation strategy compared to LH, which may be due to the fact that MV is less prone to noise and outliers. The confusion matrix of MV shows that 205 images of class A are wrongly predicted as class B while just 25 images are wrongly predicted as class C. Neighboring classes (A\&B, B\&C) are more often confused then non-neighboring classes (A\&C, C\&A). Misclassifications between neighboring classes are to a large part due to fuzzy class boundaries leading to ambiguities. The rather low number of confusions between non-neighboring classes (A\&C), however, shows that when ambiguities along the class boundaries can be neglected our approach yields a strong discriminative abilities. The overall accuracy based on MV is 65.38\%  and for LH is 64.65\%. These results are significantly higher than the random baseline for the test set of 46\% (according to the zero rule). These results show the our approach is able to extract discriminative visual patterns that are relevant to distinguish different building conditions.

\subsection{Qualitative Analysis}
\label{subsec:qualitative}

\begin{figure}[t]
\includegraphics[width=1\linewidth]{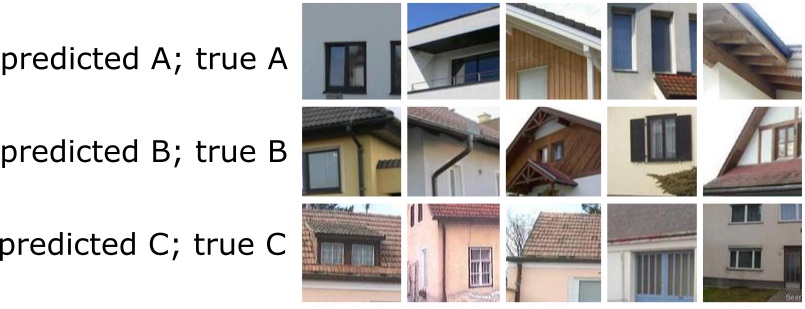}
\caption{Patches with a clear prediction for a certain building condition (likelihood > 0.99). The network successfully captures patterns related to the visually apparent condition.}
\label{fig:correct}
\end{figure}

To investigate which characteristic visual patterns are learned by the network for each building condition class, we investigate patches with different likelihoods and confidences. Figure \ref{fig:correct} shows examples of correctly predicted patches from the test set with highest likelihoods for their class (likelihood > 0.99). The patches show typical characteristics for the respective condition class, whereby the best building condition is reflected by rather newer and more modern architecture. Similarly, patches indicative for the poorest building condition belong to rather old looking and little maintained buildings. These results show that the network learns characteristic patterns that reflect the visual appearance of different building conditions. We further compute the correlation between building age and predicted condition and find a  strong positive correlation of 0.616. This is reasonable, as older buildings are more likely to exhibit a poorer condition than new buildings. Another observation is that the correctly predicted patches tend to capture rather large parts of the buildings. This shows that a certain amount of contextual information is beneficial to predict the condition. 

\begin{figure}[t]
\includegraphics[width=0.8\linewidth]{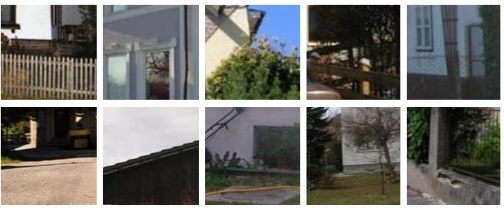}
\caption{Ambiguous patches represent less expressive structures from which the condition can hardly be estimated.}
\label{fig:ambi}
\end{figure}

Next, we select the most ambiguous patches from the test set, i.e. those patches which cannot be assigned clearly to a class and from which no information about the condition of the house can be deduced. See Figure \ref{fig:ambi} for example patches. These patches capture less expressive patches and often do not represent facade elements. Furthermore, for some patches the facade elements are occluded by trees and other objects making them less useful.

\begin{figure} [t]
\includegraphics[width=1\linewidth]{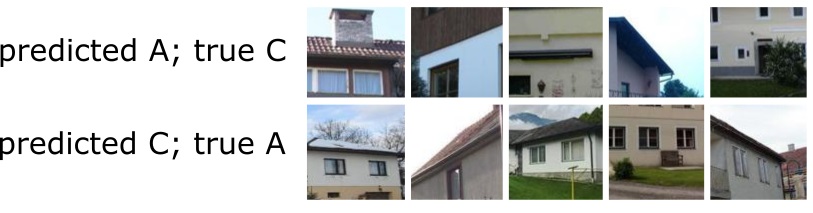}
\caption{Misclassification between non-neighboring classes (likelihood >0.99). Although predictions are wrong, they are to a great extent visually plausible which demonstrates that the visual patterns learned by the network are meaningful.}
\label{fig:incorrect}
\end{figure}

In addition to the correctly predicted and ambiguous patches, we also analyze the most confident misclassifications of our approach. Figure \ref{fig:incorrect} shows examples for the two non-neighboring and thus most distant classes (``A'' and ``C`''). From the visual examination of the patches we can understand the appraiser's assessment only to a limited extent. The prediction of the network seems to be even more plausible than the appraiser's assessment.  
There are two insights from this observation. Firstly, appraisers usually estimate the condition not only on the basis of the visual appearance of the external facade, but also on the basis of the overall impression of the building (interior space, technical equipment, etc.) which may make a difference in the assessment. Secondly, the evaluation of the condition is determined individually by each appraiser and thus, different assessments are not necessarily comparable and may live on different subjective scales. This fact becomes apparent in the context of automatic classification, for which this individual bias is  reduced due to the training from a large dataset.

\subsection{Cost Prediction of Buildings}
\label{subsec:cost prediction}
In addition to the attributes of the property, our data set also contains a value indication. All properties were valued by appraisers using the ``cost approach''. The cost approach is the standard method used in Austria to determine the market value of single-family homes. The  cost  approach  consists  of  an  estimation  of  the  land  value  plus  the  replacement cost of the building in relation to a comparable property. Replacement cost represents the estimated costs to construct (construction cost, ancillary  costs, etc.) for a new building. These costs have to be reduced in consideration of the age, condition and functional and economic obsolescence of the building by a certain \emph{discount} \cite{EVS2016}.

\begin{figure}[t]
\includegraphics[width=1\linewidth]{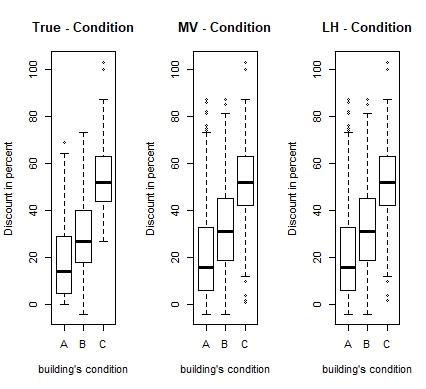}
\caption{Discount in percent of replacement cost (for new buildings) according to condition for the true condition by appraisers and the predicted conditions from MV and LH.}
\label{fig:DscountCondition}
\end{figure}

Figure \ref{fig:DscountCondition} shows the discount in percent of replacement cost (of a new building) according to condition for the true model and the predicted conditions (with majority voting, MV and average class likelihoods, LH). Both approaches are almost identical but more importantly, both models are very similar to the true assessment. This simple descriptive analysis already shows that the true model and our prediction from the images are highly correlated.

Finally, we build a regression model to predict the discount of a building and feed it with the true and the predicted condition. As an additional parameter the year of construction is added in all experiments. Table \ref{tab:regressionResult} shows the results of the regression for the three models (``True'' refers to the regression result from the manually assessed condition and ``MV'' and ``LH'' refer to the predicted conditions. In all three models we have modeled the percentage discount as a function of condition and year of construction.  
The MV and LH models are almost identical to the model built upon the appraisers' assessments. The adjusted R$^{2}$ is in all three models around 0.6, which means that about 60\% of the deviation can be explained by the models. The variable condition is coded as a dummy variable. The starting point is the state ``A'' and integrated in the intercept. Condition ``B'' has a coefficient of -0.049 in the ``True'' model (see row starting with ``true: B/A''), which means that the discount for condition B is 4.9\% higher than for condition A. In condition C, this discount is even 9.0 percent higher (see: ``true: B/A'': -0.090). This reflects well the actual relation that poorer conditions lead to larger discounts. In all models, the variable condition is significant and has the correct sign. It can be seen that the same results are obtained by image recognition and the appraiser's assessment with respect to discount. This shows that the automatically extracted condition may be a creditable substitute for the appraiser's assessments.

\begin{table}[t]
\begin{center}
    \begin{tabular}{l | l | l | l}
                  &      True &         MV &         LH      \\ \hline
  (Intercept)     &    -11.471***   & -12.132***  &  -12.149***  \\
                   &    (0.314)     &  (0.265)    &   (0.264)    \\
 year of construction          &     0.006***   &   0.006***  &    0.006***  \\ 
                   &    (0.000)     &  (0.000)    &   (0.000)    \\ \hline
  true: B/A        &    -0.049***   &             &              \\
                   &    (0.007)     &             &              \\
  true: C/A        &    -0.090***   &             &              \\
                   &    (0.011)     &             &              \\ \hline
  predictedMV: B/A &                &  -0.043***  &              \\
                   &                &  (0.007)    &              \\
  predictedMV: C/A &                &  -0.077***  &              \\
                   &                &  (0.011)    &              \\\hline
  predictedLH: B/A &                &             &   -0.045***  \\ 
                   &                &             &   (0.007)    \\
  predictedLH: C/A &                &             &   -0.077***  \\
                   &                &             &   (0.011)    \\ \hline
   adj. R$^{2}$   &     0.602      &   0.599     &    0.599     \\
  sigma            &     0.133      &   0.133     &    0.133     \\
  F                &  1021.258      & 1009.438    &  1010.674     \\
  p                &     0.000      &   0.000     &    0.000     \\ \hline 
\multicolumn{2}{l}{\scriptsize{$^{***}p<0.001$, $^{**}p<0.01$, $^*p<0.05$}}
\end{tabular}
\end{center}
    \caption{Regression results: Column two to four show the effect of the coefficients on the discount in percent of replacement cost (new building). Adj. R$^{2}$ is the adjusted coefficient of determination.}
\label{tab:regressionResult}
\end{table}

\section{Conclusion}
We have presented a first method for the estimation of building condition from unconstrained photographs. Our approach operates on a patch-level at different scales and tries to learn characteristic visual patterns related to different condition categories. Experiments yield a classification accuracy of 65\% and show that the patterns learned by the network are visually meaningful. Image-based prediction delivers equally good results in discount estimation as the assessments of the appraisers and thus predicted condition can serve as a proxy for condition estimates of appraisers. Future work will include the extension of the approach from patch-level to a global image level and to leverage network visualization techniques \cite{zhou2016learning} to identify characteristic patterns as well as multi-task learning of age and condition to exploit mutual relations. 

\section*{Acknowledgments}

We thank Sprengnetter Austria GmbH for providing real estate images and meta-data for our experiments. This work was supported by the Austrian Research Promotion Agency (FFG), Project No. 855784 and Project No. 856333. 
 

\bibliographystyle{ACM-Reference-Format}
\bibliography{sample}

\end{document}